\documentclass[sn-mathphys-num]{sn-jnl}
\usepackage{amsmath}

\usepackage{graphicx}%
\usepackage{multirow}%
\usepackage{amsmath,amssymb,amsfonts}%
\usepackage{amsthm}%
\usepackage{mathrsfs}%
\usepackage[title]{appendix}%
\usepackage{xcolor}%
\usepackage{textcomp}%
\usepackage{manyfoot}%
\usepackage{booktabs}%
\usepackage{algorithm}%
\usepackage{algorithmicx}%
\usepackage{algpseudocode}%
\usepackage{listings}%
\usepackage[utf8]{inputenc}

\theoremstyle{thmstyleone}%
\newtheorem{theorem}{Theorem}
\newtheorem{proposition}[theorem]{Proposition}%

\theoremstyle{thmstyletwo}%
\newtheorem{example}{Example}%
\newtheorem{remark}{Remark}%

\theoremstyle{thmstylethree}%
\newtheorem{definition}{Definition}%

\begin{document}

\title[Article Title]{An Enhanced Large Language Model For Cross Modal Query Understanding System Using DL-KeyBERT Based CAZSSCL-MPGPT}


\author[1]{\fnm{Shreya} \sur{Singh}}\email{shreya.singh@dituniversity.edu.in}

\affil[1]{\orgname{DIT University}, \orgaddress{\city{Dehradun}, \postcode{248009}, \state{UK}, \country{India}}}

\abstract{Large Language Models (LLMs) are advanced deep-learning models designed to understand and generate human language. They work together with models that process data like images, enabling cross-modal understanding. However, none of the existing works concentrated on avoiding the echo chamber effect in the image to enhance the accuracy of the system. Thus, the proposed system considered this limitation and developed an enhanced LLM-based framework for cross-modal query understanding using DL-KeyBERT-based CAZSSCL-MPGPT. Firstly, the collected dataset consists of images and texts. Initially, the images are pre-processed. These preprocessed images then undergo object segmentation using Easom-You Only Look Once (E-YOLO). Then, the object skeletons are generated, and a knowledge graph is constructed using a Conditional Random Knowledge Graph (CRKG). After that, features are extracted from the knowledge graph, generated skeletons, and segmented objects. The optimal features are then selected using the Fossa Optimization Algorithm (FOA). Meanwhile, the text undergoes word embedding using DL-KeyBERT. Finally, the cross-modal query understanding system utilizes CAZSSCL-MPGPT to generate accurate and contextually relevant image descriptions as text. The proposed CAZSSCL-MPGPT achieved an accuracy of 99.14187362\% in the COCO dataset 2017 and 98.43224393\% in the vqav2-val dataset.}


\maketitle

\section{Introduction}\label{sec1}

LLMs have significantly impacted Artificial Intelligence (AI) in recent years, enabling machines to understand and generate human-like text (Wang et al., 2024). These models are widely applied across various domains, including Natural Language Processing (NLP), and powering conversational agents like chatbots (Fan et al., 2024). Prominent examples include the Generative Pre-trained Transformer (GPT) family, such as ChatGPT and GPT-4, which are well-known for their strong performance and wide range of applications (Messina et al., 2021). 

Multimodal Large Language Models (MLLMs) extend the capabilities of LLMs by integrating both textual and visual data for tasks, such as cross-modal retrieval and VQA (Chen et al., 2024) (Chen et al., 2021). One key area of the MLLMs application is cross-modal queries, which involve retrieving information across different modalities, such as text and images (Liu et al., 2023) (Xu et al., 2024). For instance, in cross-modal retrieval, queries and results are represented as vectors in a common space. Also, similarity functions are used to identify relevant matches (Kalyan, 2024). Another prominent application of this approach is VQA, where an AI system answers questions about images (Lu et al., 2023). Other tasks like image captioning and image retrieval also fall under Visio-linguistic tasks that integrate visual and textual information (Zhang et al., 2021) (Chen et al., 2021a). These advancements are essential in applications, such as search engines, enabling accurate retrieval of relevant data across modalities (Liu et al., 2024).

Recent advancements in VQA and other cross-modal tasks include methods like Convolutional Neural Networks (CNNs), Long Short-Term Memory (LSTM, and Graph Convolutional Networks (GCNs) (Salaberria et al., 2023) (Ye et al., 2024). Despite these advancements, challenges like difficulty in capturing long-term dependencies and scalability issues occur when processing complex data (Tingting et al., 2023). Also, none of the existing works concentrated on avoiding the echo chamber effect that arose from the images. To mitigate this issue, a DL-KeyBERT-based CAZSSCL-MPGPT framework is proposed, enhancing performance in cross-modal tasks. 

\subsection{Problem Statement}\label{subsec1}
The existing research methodologies for cross-modal query understanding systems faced several limitations, which are outlined below,
\begin{itemize}
    \item Existing Multi-Modal Query Understanding systems, such as VQA and Image Captioning, did not address the echo chamber effect in image processing. This effect arose from over-reliance on frequent patterns in the training data, which impacted accuracy.
    \item The existing (Cheng et al., 2021) failed to capture the intricate semantic relationships and dependencies between modalities. As a result, the existing work compromised the contextual significance of each data type, limiting the improvement of the LLM.
    \item In the existing work (Zhang et al., 2024), the dynamic alignment between multiple modalities was not analyzed and unseen data was not classified correctly, limiting the improvement of the LLM. 
    \item The prevailing (Li et al., 2024) struggled with captioning highly detailed or crowded images, which often included numerous objects, scenes, or backgrounds. It faced difficulty in highlighting the most relevant features of such images.
    \item Most of the Multi-Modal Query Processing Systems used unprocessed images, leading to misleading content due to excessive noise.
\end{itemize}

\subsection{Objectives}\label{subsec2}
The objectives of the proposed work for understanding cross-modal queries are listed below.
\begin{itemize}
    \item To address the echo chamber effect in images and improve accuracy, the proposed solution involves constructing a knowledge graph using CRKG. This approach mitigates the model's overreliance on frequent patterns present in the training data.
    \item To improve LLM by capturing intricate semantic relationships and dependencies between modalities, DL-KeyBERT is utilized.
    \item The analysis of dynamic alignment between multiple modalities and the accurate classification of unseen data is achieved by introducing the cross-attention layer and zero-shot semantic consistency learning (CAZSSCL) to enhance the LLM.
    \item An E-YOLO-based object detection technique is used to better analyze complex image structures, enhancing LLM performance.
    \item The issue of misleading content caused by unprocessed images is solved by processing the image using an MF and PG-CLAHE to reduce noise and enhance image quality for improved LLM accuracy.
\end{itemize}
The remaining structure of the paper is organized as follows: Section 2 discusses the related works; Section 3 presents the proposed methodology; Section 4 reveals the results and provides a discussion; and Section 5 concludes the paper with future work.

\section{Literature Survey}\label{sec2}

(Cheng et al., 2021) deployed a deep semantic alignment network in Remote Sensing (RS) for cross-modal image-text retrieval. The Semantic Alignment Module used attention and gate mechanisms to optimize features, enhancing image-text retrieval. The model achieved improved performance in image-text matching. However, it struggled to capture intricate semantic relationships and dependencies between modalities, which compromised the contextual significance of each data type.

(Zhang et al., 2024) developed EarthGPT, a universal MLLM for multi-sensor image understanding in the RS domain. EarthGPT integrated a visual-enhanced perception mechanism to refine both coarse and fine visual information. It used cross-modal comprehension and unified instruction tuning, improving RS visual interpretation across sensor data. However, the model failed to classify unseen images that it had never encountered during the training phase. It also struggled to effectively analyze the dynamic alignment between multiple modalities, such as text and image.

(Li et al., 2024) presented an interactive perception network for LLMs. The framework extracted global and fine-grained image features using the Contrastive Language–Image Pre-training (CLIP). The Request-based Visual Information-seeking module enabled dynamic interaction, allowing LLMs to generate effective responses. Thus, the model allowed LLMs to incorporate the desired visual information for various human queries. Nevertheless, the model struggled with captioning images by highlighting the most relevant or interesting features due to the presence of multiple objects, scenes, or backgrounds presented in the crowded or detailed images.

(Lim et al., 2024) introduced a framework for Unification, Retrieval, and Generation in multimodal question answering using pre-trained language models. This framework utilized a Large Language and Vision Assistant (LLaVA) to generate detailed image descriptions. Next, the Flan-Text-to-Text Transfer Transformer (T5)-base model was fine-tuned for answer generation. The model excelled in both question-answering and retrieval tasks. However, the LLaVA model struggled with complex visual details, specialized domains, and multiple images.

(Mashrur et al., 2024) presented a VQA model using semantic Cross-Modal Augmentation (CMA). The input images were processed through CMA's mixer to create augmented replicas. Next, a back-translator was used to create multiple augmentations of the question text. Then, batched predictions were generated using the vision-language model. The model effectively handled unanswerable questions and demonstrated strong generalizability. However, the back translation failed to identify typos and punctuation issues, thus limiting its accuracy in certain cases.

(Verma et al., 2022) introduced an encoder-decoder model for automatic image caption generation. Firstly, the images were collected, and the features were extracted using the Visual Geometry Group16 Hybrid Places 1365 model as an encoder. Next, these features were fed into an LSTM-based decoder to generate captions word by word. Thus, the model produced grammatically correct captions for the input images. Nevertheless, LSTM’s limited parallelization made them less suited for large-scale or real-time applications.

(Zhu et al., 2021) developed a Multi-grained Cross-modal similarity Query with Interpretability (MCQI) framework for processing queries. Firstly, a Region Convolutional Neural Network was used to detect objects in the image. Then, LSTM and an attention mechanism identified latent semantic relationships. When a query was executed, the MCQI framework processed it through the index using refined k-Nearest Neighbors. This approach improved the scalability of the MCQI framework. However, the model was heavily dependent on large training data from specific areas, limiting its generalization.

(Dong et al., 2021) illustrated cross-modal retrieval using an Adversarial GCN (AGCN). The model employed a graph feature generator based on a GCN. It utilized a minimax game strategy with a graph feature discriminator to ensure modality-invariant feature representations. The results showed that AGCN improved cross-modal retrieval accuracy. However, the minimax game strategy in the AGCN model introduced significant computational complexity, making it resource-intensive and less efficient.

(Sun et al., 2021) demonstrated a Cross-Modal Pre-Aligned method with Global and Local information (CMPAGL) for efficient RS image-text retrieval. The model used a Gswin transformer block for feature extraction. Then, text features were embedded using a Bidirectional Encoder Representations from Transformers (BERT)-based encoder. The model showed superior performance in retrieving image-text pairs. However, BERT was slow to train due to its large size and the extensive number of weights that needed updating.

(Zhang et al., 2024a) explored an enhanced feature extraction framework to enhance cross-modal image-text retrieval. The Enhanced Vision Transformer extracted the image features. Then, a triplet loss function optimized the model. The framework achieved high detection accuracy. However, it struggled to balance dynamic adaptation to both large and small feature extraction requirements, affecting model efficiency and generalization.

\section{Proposed Methodology For Cross-Modal Query Understanding System Using Enhanced LLM}
This research paper proposes an advanced LLM for understanding cross-model queries using the DL-KeyBERT-based CAZSSCL-MPGPT technique. The main phases of the proposed work are: dataset collection, image preprocessing, object segmentation, object skeleton generation, knowledge graph construction, text word embedding, feature extraction, feature selection, and the cross-modal query understanding system. The structural diagram of the proposed framework is illustrated in Figure 1.

\begin{figure}
    \centering
    \includegraphics[width=1\linewidth]{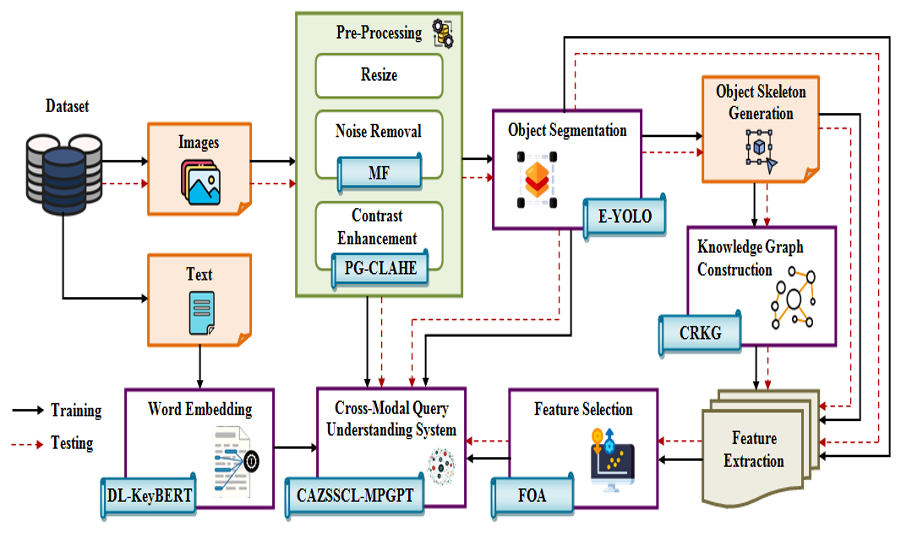}
    \caption{Structural Diagram of the Proposed Framework}
    \label{fig:enter-label}
\end{figure}

\subsection{Input Dataset}
The input data is collected from a public dataset consisting of images and corresponding textual descriptions. The texts (A) and images (B) in the dataset can be expressed as,
\begin{equation}
 A = \{ A^1, A^2, A^3, A^4, A^5, \ldots, A^{a-1}, A^a \}
\label{eq:1}
\end{equation}

\begin{equation}
 B = \{ B^1, B^2, B^3, B^4, B^5, \ldots, B^{b-1}, B^b \}
\label{eq:2}
\end{equation}
Where, (\textit{a}) and (\textit{b}) illustrate the total number of (A) and (B). 

\subsection{Pre-Processing}
In this phase, (B) is pre-processed to improve the quality of the cross-modal query understanding system. The pre-processing steps are as follows,

\subsubsection{Resize}
In this step, (B) are resized to a uniform dimension to ensure consistency for further processing. The resized image ($\Re$) is represented as
\begin{equation}
\Re=\psi(B)
\label{eq:placeholder_label}
\end{equation}
Where, ($\psi$) indicates the resizing function. 

\subsubsection{Noise Removal}
Next, the unwanted noises in ($\Re$) are removed using MF, which eliminates random noise that could degrade data quality. Thus, only relevant visual content is retained. At first, a window ($\omega$) is selected for each pixel and moves across the image to process it pixel by pixel. Next, within ($\omega$), the pixel values are collected and then sorted in ascending order. The sorted pixels are denoted as ($\Re$'), and the median value (M) is calculated by,

\begin{equation}
    M = 
    \begin{cases} 
        \mathfrak{R'}_{(\frac{n+1}{2})}, & \text{if } n \text{ is odd} \\ 
        \frac{\mathfrak{R'}_{(\frac{n}{2})} + \mathfrak{R'}_{(\frac{n+1}{2})}}{2}, & \text{if } n \text{ is even}
    \end{cases}
    \label{eq:placeholder_label}
\end{equation}

Where, (\textit{n}) depicts the number of pixel values present in ($\Re$'). Next, the central pixel (C) is replaced with the calculated (M). It is formulated as,
\begin{equation}
C = M
\label{eq:placeholder}
\end{equation}

By replacing each pixel with (M) from its neighborhood, the random noise is effectively removed. Finally, the noise-removed image is represented as (N).

\subsubsection{Contrast Enhancement}
In this step, the contrast of (N) is enhanced using PG-CLAHE to make the relevant features more distinguishable and to enhance the visual quality of the image. Contrast Limited Adaptive Histogram Equalization (CLAHE) improves local contrast in areas with varying intensities. It also avoids over-enhancing the noise in uniform or homogeneous regions. However, CLAHE uses a clip limit parameter, and an improper setting of this parameter can result in over-enhancement of the image. To address this issue, the Pareto Gini distribution technique is utilized to determine the optimal clip limit for the CLAHE algorithm. The steps involved in PG-CLAHE are described below,

Initially, (N) is divided into small, non-overlapping tiles (G) for local processing and region-specific enhancement. Next, for each (G), histogram equalization is performed to enhance the contrast. The histogram ($\xi$) of each tile is calculated as,

\begin{equation}
        \xi = \sum_{(x,y) \in G} G(x,y)
    \label{eq:placeholder_label}
\end{equation}

Where, \textit{G(x,y)}  denotes the pixel intensity value at co-ordinates \textit{(x,y)} within the \textit{(G)}. Next, a clipping limit ($\chi$) is applied to prevent excessive contrast enhancement. The Pareto Gini distribution technique is used to determine the optimal clipping limit. 
\begin{equation}
        \chi \xrightarrow{} \widetilde{\chi}\ = 1- \frac{2}{\sigma + 1}
    \label{eq:placeholder_label}
\end{equation}

Here, ($\sigma$)  signifies the Pareto exponent and ($\widetilde{\chi}$)  denotes the Gini index. The clipped image (H) is represented as,
\begin{equation}
        H=min(\xi,\widetilde{\chi})
    \label{eq:placeholder_label}
\end{equation}
For each (H), the Cumulative Distribution Function (CDF) is calculated to map the intensity values to the desired output image. The CDF ($H'''$) is equated as,
\begin{equation}
        H'''=\frac{\sum_{G=0}^{g} H(G)}{\iota_{pix}}
    \label{eq:placeholder_label}
\end{equation}
Where, ($\iota_{pix}$)  represents the total number of pixels in (H) and (\textit{g}) indicates the total number of tiles. Next, to enhance the dynamic range and maximize contrast, the CDF is normalized. The normalized CDF ($\mu(x,y)$)  is formulated as,
\begin{equation}
        \mu(x,y)=\frac{H'''-min(H''')}{max(H''')-min(H''')}
    \label{eq:placeholder_label}
\end{equation}
Where, ($min(H''')$) and ($max(H''')$) denotes the minimum and maximum of ($H'''$), respectively. To prevent boundary artifacts and ensure smooth transitions between adjacent tiles, interpolation is applied. The interpolated pixel value ($\breve{\mu}$)  is computed as,
\begin{equation}
    \breve{\mu}=(1-p)(1-q)\mu(x,y)+p(1-q)\mu(x+1,y)+pq\mu(x+1,y+1)+(1-p)q\mu(x,y+1)
    \label{eq:placeholder_label}
\end{equation}
Here, (\textit{p}) and (\textit{q}) illustrates the fractional horizontal and vertical distances between two adjacent tiles, respectively. Thus, the contrast-enhanced image is specified as ($E'$). The pseudocode for the proposed PG-CLAHE is illustrated as, \break

\hrule
\vspace{0.2cm}
\noindent
\textbf{\large Pseudocode for PG-CLAHE}

\vspace{0.2cm}
\noindent
\textbf{Input:} Noise Removed Image (\(N\))\\
\textbf{Output:} Contrast Enhanced Image (\(E'\))

\hrule
\vspace{0.2cm}

\noindent
\textbf{Begin}

\quad \textbf{Initialize} (\(G\)), (\(x,y\)), (\(p\)), (\(q\)), (\(\sigma\)), (\(\widetilde{\chi}\))

\quad \textbf{For each} (\(N\)) \textbf{do}

\quad\quad \textbf{Divide} (\(N\)) \(\xrightarrow{to} G\)

\quad\quad \textbf{Calculate} histogram (\(\xi\)) \textbf{for each} (\(G\))

\quad\quad \textbf{Compute} (\(\chi\))

\quad\quad\quad $\chi \xrightarrow{} \widetilde{\chi}\ = 1- \frac{2}{\sigma + 1}$

\quad\quad \textbf{Select} clipping limit (\(\chi\))

\quad\quad \textbf{Obtain} (H) by applying (\(\chi\))

\quad\quad\quad $H=min(\xi,\widetilde{\chi})$

\quad\quad \textbf{Evaluate} ($H'''$) \#CDF

\quad\quad\quad $H'''=\frac{\sum_{G=0}^{g} H(G)}{\iota_{pix}}$

\quad\quad \textbf{Normalize} the ($H'''$)

\quad\quad\quad $\mu(x,y)=\frac{H'''-min(H''')}{max(H''')-min(H''')}$

\quad\quad \textbf{Compute} ($\breve{\mu}$)

\quad \textbf{End} for

\quad \textbf{Return} ($E'$)

\textbf{End} \break

\hrule
\vspace{0.2cm}

\noindent
Finally, the pre-processed image is denoted as ($\zeta$).

\subsection{Object Segmentation}
In this phase, the objects in ($\zeta$) are segmented using E-YOLO to differentiate various objects, scenes, or backgrounds present in the image. Here, You Only Look Once (YOLO) is utilized for its high speed and accuracy in identifying objects. YOLO provides an accurate bounding box for each detected object and enhances object localization. However, YOLO struggles to detect objects that are far from the camera or small in size. To address this limitation, the Easom function is used to find the scaling factor. This function is also integrated with overlapping elimination to effectively identify small objects, thereby improving the efficiency of localization. At first, ($\zeta$) is divided into a number of grids ($Q$) for detecting the objects that fall within its boundaries. Next, for each ($Q$), ($r$) numbers of bounding boxes ($\beta$) are predicted with their confidence score ($\ddot{c}^r$). 
 
\begin{equation}
    \beta=[o^r,\dddot{o}^r,\tau^r,\dddot{\tau}^r,\ddot{c}^r]^{r=1 \ to\ \widetilde{r}}
    \label{eq:placeholder_label}
\end{equation}
Where, ($o^r$)  and  ($\dddot{o}^r$) refers to the coordinates of the centre of the bounding box relative to the grid cell, respectively,   ($\widetilde{r}$) denotes the total number of ($r$), and ($\tau^r$)  and ($\dddot{\tau}^r$) indicate height and width of ($\beta$), respectively. Then, the Easom function ($\zeta$) is employed to identify a scaling factor for ($\beta$).
\begin{equation}
    \zeta=-cos(i) \cdot cos(j) \cdot exp(-((i-\pi)^2+ (j-\pi)^2))
    \label{eq:placeholder_label}
\end{equation}
Here, ($\pi$) indicates the mathematical constant and ($i$) and ($j$) represent the dimensions of ($\beta$). Thus, the scaling factor ($S$) is identified from ($\zeta$), and ($\beta$) is adjusted.
\begin{equation}
    \beta'=\beta * \zeta
    \label{eq:placeholder_label}
\end{equation}
Where, ($\beta'$) indicates the adjusted ($\beta$). Further, class probabilities ($P$) are predicted for each object-detected image. 
\begin{equation}
    P=\delta(\ddot{c}_{h'''} | \beta')
    \label{eq:placeholder_label}
\end{equation}
\begin{equation}
    \delta(\beta')=\frac{e^{\beta'_{y''}}}{\sum_{y'''}^{\widetilde{j}} e^{\beta'_{y'''}}}
    \label{eq:placeholder_label}
\end{equation}
Here, ($\ddot{c}_{h'''}$)  represents the class scores, which is the raw value indicating class likelihood, ($\delta$) indicates the softmax activation function, ($e$)  refers to the Euler’s value, ($\beta'_{y''}$)  indicates the ($y''^{th}$)  class of ($\beta'$), ($y'''=1 \ to \ \widetilde{j}$)  represents the sum of the exponentials of all class scores, ($\widetilde{j}$)  refers to the total number of ($y'''$), and ($\beta'_{y'''}$) signifies the ($y'''^{th}$)  index of ($\beta'$). Now, Non-Maximum Suppression is applied to eliminate redundant overlapping bounding boxes by calculating the Intersection Over Union (IOU) for each ($\beta'$), thus attaining final segmented objects ($\vartheta$).
\begin{equation}
    \vartheta=P*\kappa*\zeta
    \label{eq:placeholder_label}
\end{equation}
\begin{equation}
    \kappa=\frac{{|\beta^d \cap \beta^{\overleftrightarrow{d}|}}}{|\beta^d \cup \beta^{\overleftrightarrow{d}}|} 
    \label{eq:placeholder_label}
\end{equation}
Where, ($\kappa$) indicates the IOU, ($\beta^d$) and ($\beta^{\overleftrightarrow{d}}$) illustrate ($d^{th}$) and ($\overleftrightarrow{d}^{th}$) bounding boxes, correspondingly, and ($\cap$) and ($\cup$) signifies the intersection and union operations, respectively. Thus, ($\vartheta$) are obtained.

\subsection{Object Skeleton Generation}
Next, the object skeletons are generated from ($\vartheta$) to capture the core structure of each object. This simplifies the representation by focusing on the object's shape and removing unnecessary details. The skeletonized images ($\rho$) are illustrated as,
\begin{equation}
    \rho=\{\rho_1,\rho_2,\rho_3,\rho_4,......\rho_l\}
    \label{eq:placeholder_label}
\end{equation}
Where, ($l$) refers to the total number of ($\rho$).

\subsection{Knowledge Graph Construction}
Here, a knowledge graph is constructed from ($\rho$) using CRKG. This graph is designed to address the echo chamber effect in images and improve accuracy. The echo chamber effect arises when the model relies heavily on frequent patterns in its training data, leading to biased results. The Knowledge Graph (KG) integrates diverse data sources and provides a comprehensive view of the information. By representing entities and their relationships, the knowledge graph enables machines to understand the context of the data. However, KG can also perpetuate biases present in the underlying data and struggle to capture dynamic relationships in the images. To mitigate this issue, the conditional random technique is used. This technique reduces the echo chamber effect by accounting for dependencies between entities and relationships, ensuring a more balanced and accurate graph.

Firstly, the entities ($J^{en}$) are identified from ($\rho$)  by considering their context and spatial relationships between regions. Here, the conditional random technique is used to label the entities. 
\begin{equation}
    R(Y|\rho)=\frac{1}{\overleftrightarrow{R}(\rho)} exp(\sum_{l}w^lF^l(Y,\rho))
    \label{eq:placeholder_label}
\end{equation}
Here, ($R(Y|\rho)$)  denotes the conditional probability,   ($\overleftrightarrow{R}(\rho)$) refers normalization factor,  ($w^l$) signifies the weight parameter associated with a specific feature of ($l$), and ($F^l(Y,\rho)$) illustrates the feature function that computes the relationship between input ($\rho$) and output sequences ($Y$), which represents labelled entities derived from ($\rho$). Next, the relationships ($J^{re}$) between these  ($J^{en}$) are identified. 
\begin{equation}
    J^{re}=\{\overleftrightarrow{j} | J'^{en} \ is \ related \ to \ J''^{en}\}
    \label{eq:placeholder_label}
\end{equation}
Here, ($\overleftrightarrow{j}$)  refers to the nature of their relationship and ($J'^{en}$) and  ($J''^{en}$) indicates the entities in the graph. Next, the knowledge graph ($K$) is constructed by representing entities as nodes and relationships as edges.
\begin{equation}
    K=\langle J^{en},J^{re}\rangle
    \label{eq:placeholder_label}
\end{equation}
Thus, ($K$)  is then given as input to the feature extraction phase for further processing.

\subsection{Feature Extraction}
Now, various image features, such as color, edge, texture, Gray-Level Co-occurrence Matrix (GLCM), Histogram of Oriented Gradients (HOG), mean, variance, kurtosis, skewness, smoothness, correlation, etc., are extracted from ($\vartheta$) and ($\rho$) . Similarly, from ($K$), features, such as node attributes, edge relationships, node degree, closeness centrality, pagerank, etc., are also extracted. These extracted features ($E$) are represented as,
\begin{equation}
    E=\{E_1,E_2,E_3,E_4,......E_m\}
    \label{eq:placeholder_label}
\end{equation}
Where,  ($m$) represents the total numbers of ($E$).

\subsection{Feature Selection}
Then, optimal features are selected from ($E$)  using FOA. FOA is used to identify the optimal subset of features for enhancing the model performance by efficiently navigating the solution space. It balances exploration and exploitation, which helps to avoid local optima. Firstly, the population matrix ($Z$)  represents that the initial positions of the fossa are randomly initialized within the feature space based on ($E$). The ($Z$) is defined by,
\begin{equation}
Z=
    \begin{bmatrix}
    Z_1 \\
    \vdots \\
    Z_k \\
    \vdots \\
    Z_D
\end{bmatrix}
=
    \begin{bmatrix}
    z_{1,1}  & \hdots z_{1,t} & \hdots & z_{1,u} \\
    \vdots & \ \ \vdots & \ \ \ \ \  \vdots \\
    z_{k,1}  & \hdots z_{k,t} & \hdots & z_{k,u} \\
    \vdots & \ \ \vdots & \ \ \ \ \  \vdots \\
    z_{D,1}  & \hdots z_{D,t} & \hdots & z_{D,u} \\
\end{bmatrix}
    \label{eq:placeholder_label}
\end{equation}
\begin{equation}
    z_{k,t}=(ub_t -lb_t)*\varphi+lb_t
    \label{eq:placeholder_label}
\end{equation}
Where, ($Z_k$)  denotes the ($k^{th}$)  fossa, ($z_{k,t}$)  illustrates the ($t^{th}$)  dimension of ($k^{th}$)  fossa in the search space, ($u$)  refers to the number of decision variables,  ($D$) indicates the number of fossa, ($\varphi$)  signifies a random number, and ($ub_t$)  and ($lb_t$)  determines the upper and lower bound of ($t^{th}$) dimension variable, respectively. Next, the fitness ($\Phi_h$) of each fossa is evaluated based on the maximum classification accuracy ($max(\hbar^{class})$). 
\begin{equation}
    \Phi_h=max(\hbar^{class})
    \label{eq:placeholder_label}
\end{equation}
After that, in the exploration phase, the candidate lemurs ($\delta_k$)  (better solutions) for each fossa are determined as,
\begin{equation}
    \delta_k=\{Z_h:\Phi_h < \Phi_k \ and \ h \neq k\},where \ k=1,2,..,D\ and \ h \in \{1,2,..,D\}
    \label{eq:placeholder_label}
\end{equation}
Where, ($Z_h$)  expresses the position of the ($h^{th}$) fossa with a better ($\Phi_h$). Here, ($h$)  and  ($k$) denotes the index of the current fossa and selected fossa (lemur), respectively. Next, the fossa randomly selects a lemur ($\lambda_{k,t}$) from ($\delta_h$)  and adjusts its position as,
\begin{equation}
    z_{k,t}^{\hat{P}1}=z_{k,t} + \varphi_{k,t}*(\lambda_{k,t}-I_{k,t}*z_{k,t})
    \label{eq:placeholder_label}
\end{equation}
Where, ($Z_{k}^{\hat{P}1}$)  signifies the new position of the ($k^{th}$)  fossa during the attack, ($\Phi_{k,t}$)  illustrates the random values, and ($I_{k,t}$)  indicates the random integer. If the new position  ($Z_{k}^{\hat{P}1}$) improves the objective function, it replaces the current position as,
\begin{equation}
    Z_k=
    \begin{cases} 
    Z_{k}^{\hat{P}1},\ \ \Phi_{k}^{\hat{P}1} < \Phi_k \\
    Z_k, \  \ else
    \end{cases}
    \label{eq:placeholder_label}
\end{equation}
Where, ($\Phi_{k}^{\hat{P}1}$)  determines the objective function value for the new position. Next, in the exploitation phase, the fossa chases the lemur. The fossa adjusts its position to mimic the pursuit, which is given by,
\begin{equation}
    z_{k,t}^{\hat{P}2}=z_{k,t} + (1-2\varphi_{k,t})*\frac{ub_t-lb_t}{t''}
    \label{eq:placeholder_label}
\end{equation}
Here, ($t''$)  denotes the iteration count and ($Z_{k,t}^{\hat{P}2}$)  illustrates the updated position for the  ($k^{th}$) fossa. If the new position ($Z_{k}^{\hat{P}2}$)  improves the objective function, it replaces the current position as,
\begin{equation}
    Z_k=
    \begin{cases} 
    Z_{k}^{\hat{P}2},\ \ \Phi_{k}^{\hat{P}2} < \Phi_k \\
    Z_k, \  \ else
    \end{cases}
    \label{eq:placeholder_label}
\end{equation}
Where,  ($\Phi_{k}^{\hat{P}2}$) determines the objective function value for the updated position. Thus, after going through several iterations, the optimal features ($O$) are selected.

\subsection{Word Embedding}
In this phase, word embedding is performed for the collected texts  ($A$) using DL-KeyBERT to convert text data into a vector format. This embedding technique captures intricate semantic relationships and dependencies between modalities, ultimately improving the performance of the LLM. KeyBidirectional Encoder Representations from Transformers (KeyBERT) is a minimal and easy-to-use embedding technique that leverages BERT embeddings. It transforms text into vector representations that capture the semantic meaning of words and their context. However, KeyBERT’s embeddings are based on cosine similarity between word and document embeddings, which does not always guarantee relevance. Also, it may produce embeddings that are semantically close but contextually meaningless. To address this, Damerau-Levenshtein (DL), a method for orthographic similarity, is used to identify the top (\u{n})  most similar terms in KeyBERT for improved word embedding quality. Firstly, ($A$)  is divided into small units called tokens ($\widetilde{A}$). Next, an embedding vector of ($\widetilde{A}$)  is generated using a pre-trained model i.e., BERT. The embedded vector ($\dddot{E}$) is equated as,
\begin{equation}
    \dddot{E}=\Im(\widetilde{A})
    \label{eq:placeholder_label}
\end{equation}
Here, ($\Im$)  denotes the transformer model. The candidate keywords ($X$) are then extracted from ($E$). Next, candidate keyword embeddings ($\widetilde{X}$) are generated as,
\begin{equation}
    \widetilde{X}=\Im(X)
    \label{eq:placeholder_label}
\end{equation}
Then, the DL distance (L(\u{a},\newtie{b})) is computed between ($\widetilde{A}$)  and ($\widetilde{X}$)  to identify the top  (\u{n}) similar terms. It is computed as,
\begin{equation}
    L(\breve{a},\text{\newtie{b}})=
    \begin{cases} 
    max(\breve{a},\text{\newtie{b}}), \qquad\qquad\qquad\qquad\qquad if \ min(\breve{a},\text{\newtie{b}})=0 \\
    min(L(\breve{a}-1,\text{\newtie{b}})+1), \qquad\qquad\qquad\qquad (del) \\
    min(L(\breve{a},\text{\newtie{b}}-1)+1), \qquad\qquad\qquad\qquad (ins) \\
    min(L(\breve{a}-1,\text{\newtie{b}}-1)+c^{sub}(\breve{a},\text{\newtie{b}})), \qquad\quad (sub) \\
    min(L(\breve{a}-2,\text{\newtie{b}}-2)+c_{tra}(\breve{a},\text{\newtie{b}})), \qquad\quad (tra) \\
    \end{cases}
    \label{eq:placeholder_label}
\end{equation}
Here, (L(\u{a},\newtie{b}))  refers to the distance between the first (\u{a})  and  (\newtie{a}) characters of string  ($\widetilde{A}$)  and ($\widetilde{X}$), respectively, $c^{sub}(\breve{a},\text{\newtie{b}})$  and  $c_{tra}(\breve{a},\text{\newtie{b}})$ denotes substitution and transposition cost, correspondingly, and $(del)$ ,$(ins)$ ,$(sub)$ , and  $(tra)$ denotes deletion, insertion, substitution, and transposition operations, respectively. Next, ($\widetilde{X}$)  are ranked in the input text embedding based on (L(\u{a},\newtie{b})). Finally, the word-embedded text ($\Xi$) is obtained by extracting the top  (\u{n}) most relevant keywords ($\Xi$).
\begin{equation}
    \Xi=\langle{\Xi_1,\Xi_2,\Xi_3,\Xi_4,......\Xi_f}\rangle    \label{eq:placeholder_label}
\end{equation}
Here, ($f$)  denotes the total number of ($\Xi$). The pseudocode for the DL-KeyBERT is described below, \break

\hrule
\vspace{0.2cm}
\noindent
\textbf{\large Pseudocode for DL-KeyBERT}

\vspace{0.2cm}
\noindent
\textbf{Input:} Collected Text (A) \\
\textbf{Output:} Word Embedded Text (\(\Xi\))

\hrule
\vspace{0.2cm}

\noindent
\textbf{Begin}

\quad \textbf{Initialize} $(\Im)$, \u{a}, \newtie{b}

\quad \textbf{For each} (\(A\)) \textbf{do}

\quad\quad \textbf{Tokenize} (\(A\)) \ to \ ($\widetilde{A}$)

\quad\quad \textbf{Compute} embedded \ vector \  ($\dddot{E}$) \ \#by \ using \ BERT 

\quad\quad \textbf{Extract} (\(X\)) \ from \ ($\dddot{E}$)

\quad\quad \textbf{Generate} ($\widetilde{X}$)

\quad\quad\quad     $\widetilde{X}=\Im(X)$

\quad\quad \textbf{Calculate}\ DL \ distance \ L(\u{a}, \newtie{b})

\quad\quad\quad \textbf{If} max(\u{a}, \newtie{b})

\quad\quad\quad\quad \textbf{Compute} min(\u{a}, \newtie{b})=0

\quad\quad\quad \textbf{Else} 

\quad\quad\quad\quad \textbf{Compute} 

\quad\quad\quad \textbf{End} \ if

\quad\quad \textbf{Rank}\ the \ ($\widetilde{X}$) \ based \ on \ L(\u{a}, \newtie{b})

\quad\quad \textbf{Select}\ top \ keywords

\quad\quad \textbf{Obtain}\ vector \ values \ ($\Xi$)

\quad\quad\quad $\Xi=\langle{\Xi_1,\Xi_2,\Xi_3,\Xi_4,......\Xi_f}\rangle$

\quad \textbf{End} for

\quad \textbf{Return} ($\Xi$)

\noindent
\textbf{End} \break

\hrule
\vspace{0.2cm}

This output is further given as input to the cross-modal query understanding system for training the model to perform well on real-time unseen image caption generation.

\subsection{Cross-Modal Query Understanding System}

In this phase, the CAZSSCL-MPGPT model is trained using ($\vartheta$), ($O$), ($\Xi$), and ($\zeta$)  to generate image captions. The attention-based design of GPT helps the model keep track of context across different inputs, allowing it to understand complex queries that involve multiple types of data. GPT models are trained on massive datasets, which can inadvertently contain real-world biases. This can lead to the generation of biased or discriminatory outputs. To address this, a Mixup regularization scheme is used, thus selecting relevant features and reducing model complexity and bias. Further, the GPT model uses the GELU activation function; but, it is computationally complex and less efficient. To improve efficiency, the Phish activation function is used, thus maintaining linearity for positive inputs and slight non-linearity for negative inputs. The model also incorporates a cross-attention layer and Zero-Shot Learning (ZSL) to analyze the dynamic alignment between multiple modalities and correctly classify unseen data. ZSL enables a model to recognize objects or generate outputs for categories not seen during training. However, ZSL can struggle with semantic gaps between seen and unseen classes, leading to failures in recognizing or classifying unseen classes if their descriptions differ too much from the seen ones. To address this, Semantic Consistency Loss is used, thus aligning the semantic representations of seen and unseen classes for better recognition and classification accuracy. The diagrammatic representation of the proposed CAZSSCL-MPGPT is illustrated in Figure 2.

\begin{figure}
    \centering
    \includegraphics[width=1\linewidth]{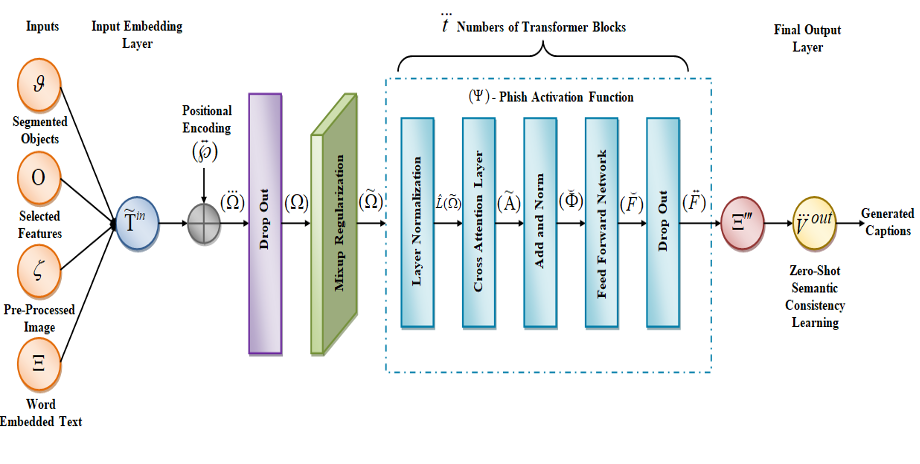}
    \caption{Diagrammatic Representation of the Proposed CAZSSCL-MPGPT}
    \label{fig:enter-label}
\end{figure}

\textbf{Input Embedding Layer:}
In the training time, initially, ($\vartheta$), ($O$), ($\Xi$), and ($\zeta$) are given as input to the CAZSSCL-MPGPT model and are commonly denoted as (${T}^{in}$). Then, (${T}^{in}$) is transformed into embeddings ($\widetilde{T}^{in}$), and positional encoding ($\overleftrightarrow{\wp}$) is added to provide token order information. The output of the embedding layer  ($\dddot{\Omega}$)  is equated as,
\begin{equation}
    \dddot{\Omega}=(\widetilde{T}^{in}) \oplus \overleftrightarrow{\wp}
    \label{eq:placeholder_label}
\end{equation}
Where, ($\dddot{\Omega}$) has ($o'$)  number of inputs.  Then, dropout is applied to ($\dddot{\Omega}$)  to prevent overfitting, and the output is denoted as ($\Omega$).

\textbf{Mixup Regularization}
The mixup regularization is used to select only the relevant inputs and neglect the irrelevant ones. It is illustrated as,
\begin{equation}
    \widetilde{\Omega}=\upsilon\Omega_{o'}+(1-\upsilon)\Omega_{o'+1}
    \label{eq:placeholder_label}
\end{equation}
\begin{equation}
    \Theta=\upsilon\Theta_{o'}+(1-\upsilon)\Theta_{o'+1}
    \label{eq:placeholder_label}
\end{equation}
Where, ($\widetilde{\Omega}$)  refers to the new input created by mixing the original inputs ($\Omega_{o'}$)  and ($\Omega_{o'+1}$),  ($\Theta$)  illustrates the new label created by mixing the original labels ($\Theta_{o'}$)  and  ($\Theta_{o'+1}$), and ($\upsilon$)  indicates random scalar taken from a Beta distribution, which is the mixing coefficient.

\textbf{Transformer Blocks}
After mixup regularization, the  ($\dddot{t}$) numbers of transformer blocks are computed. They are,

\textit{Layer Normalization}

Next, layer normalization ($\hat{L}(\widetilde{\Omega})$)  is employed to ensure that the output of each layer has a stable distribution, helping the model converge faster.
\begin{equation}
    \hat{L}(\widetilde{\Omega})=\frac{\widetilde{\Omega}-\dddot{\mu}}{\sqrt{\sigma^{m^2} + \epsilon}}
    \label{eq:placeholder_label}
\end{equation}
Here, ($\dddot{\mu}$), ($\sigma^{m^2}$), and ($\epsilon$)  indicates the mean, variance, and standard deviation of ($\widetilde{\Omega}$), correspondingly.

\textit{Cross Attention Layer}

The Cross-Attention Mechanism ($\widetilde{A}$) is crucial in aligning the text features with the image features by computing attention between the visual features (queries ($U$)) and text features (keys ($W$) and values ($\hat{U}$)) derived from ($\hat{L}(\widetilde{\Omega})$).
\begin{equation}
    \widetilde{A}(U,W,\hat{U})=\delta(\frac{UW^\Gamma}{\sqrt{d''_W}})\hat{U}
    \label{eq:placeholder_label}
\end{equation}
Where, ($\Gamma$)  denotes transpose, and ($d''_W$) illustrates the dimensionality keys.

\textit{Add and Norm}
Next, a residual connection is added, followed by layer normalization, to stabilize training and maintain the gradient flow. The normalized output is denoted as ($\breve{\Theta}$).

\textit{Feed Forward Network}
Here, the phish activation function ($\Psi$)  is used to solve the vanishing gradient problems. The feed-forward network output ($\breve{F}$) is given by,
\begin{equation}
    \breve{F}=\Psi(\breve{\Theta}*W^1+\hat{B}^2)*W^2+\hat{B}^2
    \label{eq:placeholder_label}
\end{equation}
\begin{equation}
    \Psi=T^{in}*tanh(GELU(T^{in}))
    \label{eq:placeholder_label}
\end{equation}
Here, ($W^1$) and ($W^2$) indicates the weight matrices, ($\hat{B}^1$) and ($\hat{B}^2$)  illustrates the bias terms, and ($tanh$) and ($GELU$) refers to the hyperbolic tangent and GELU activation functions, respectively.

\textit{Dropout}

Further, dropout is applied to ($\breve{F}$) to prevent overfitting, and the output is notated as ($\overleftrightarrow{F}$).

\textbf{Final Output Layer:}
After completing all the transformer layers, the output undergoes final layer normalization. Then, the output is passed through a linear layer to map it to the size of the vocabulary. Finally, a softmax layer is applied to generate the caption for the image. 
\begin{equation}
    \Xi'''=\delta(Li(\hat{L}(\overleftrightarrow{F})))
    \label{eq:placeholder_label}
\end{equation}
Where, ($\Xi'''$)  indicates the generated captions, ($Li$)  denotes the linear function, and ($\hat{L}(\overleftrightarrow{F})$) refers to the layer normalization performed on ($\overleftrightarrow{F}$).

\textbf{Zero-Shot Semantic Consistency Learning:}
To align the semantic representations of seen and unseen classes and to enable the model to recognize unseen classes, Zero-Shot Semantic Consistency Learning  ($\dddot{V}^{out}$) is utilized. It is given by,
\begin{equation}
    \dddot{V}^{out}=arg \ max \ \text{\newtie{P}}(\ddot{V}|\Xi''')-l_{loss}
    \label{eq:placeholder_label}
\end{equation}
\begin{equation}
    l_{loss}=\frac{1}{\hat{n}}\sum_{\hat{r}=1}^{\hat{n}}(||\widetilde{G}(\Xi'''_{\hat{r}})-\widetilde{T}(\Xi'''_{\hat{r}})||_2^2 + ||\widetilde{T}'''(\Xi'''_{\hat{r}})-\widetilde{T}(\Xi'''_{\hat{r}})||_2^2)
    \label{eq:placeholder_label}
\end{equation}
Where, ($l_{loss}$)  indicates semantic consistency loss, ($\text{\newtie{P}}(\ddot{V}|\Xi''')$)  refers to the probability distribution for the predicted class based on the input, ($\hat{n}$)  represents a total number of batch size, ($\hat{r}$) indicates batch size, ($\widetilde{G}(\Xi'''_{\hat{r}})$)  signifies generated output, ($\widetilde{T}(\Xi'''_{\hat{r}})$)  indicates target representation of the input, and ($\widetilde{T}'''(\Xi'''_{\hat{r}})$)  determines another target representation. Based on this loss function, the final ($\Xi'''$)  is generated. 

\subsection{Testing}
During the testing phase, the images are first pre-processed. Then, objects are segmented using E-YOLO, followed by the generation of object skeletons and the construction of a knowledge graph using CRKG. Afterward, features are extracted from the constructed knowledge graph, generated skeleton objects, and segmented objects. Next, optimal features are selected using FOA. Finally, the trained CAZSSCL-MPGPT model generates the image captions. Similarly, in the VQA process, an image and a question are provided as input. The image undergoes pre-processing, segmentation, and feature selection, and the question is converted into word-embedded text. In this process, for each question, the model provides the correct answer by utilizing the segmented objects, pre-processed images, and optimal features as inputs. Thus, the VQA process accurately generates answers based on the image and question. The performance evaluation of the proposed model is described in the following section.

\section{Results And Discussions}
In this section, the performance of the proposed work is evaluated and compared with existing models to demonstrate its effectiveness. The implementation is carried out using Python.

\subsection{Dataset Description}
The proposed work utilizes the Common Objects in Context (COCO) Dataset 2017 and the visual question answering v2 validation (vqav2-val) dataset to evaluate the efficiency of the proposed model. Both datasets are publicly available and are mentioned in the reference section. The COCO Dataset 2017 is a large-scale dataset, containing 328K images. The vqav2-val dataset includes both text and image data for the VQA task. Here, 80\% of the data is used for training the cross-modal query understanding system, while 20\% is reserved for testing its efficacy.

\subsection{Performance Evaluation}
This subsection compares the performance of the proposed techniques with the existing techniques, highlighting their effectiveness based on key performance metrics.

\subsubsection{Performance Analysis of Contrast Enhancement}
The proposed PG-CLAHE’s performance is analyzed and compared with the existing techniques like CLAHE, Adaptive Histogram Equalization (AHE), Histogram Equalization (HE), and Bilateral Filtering (BF).
\begin{figure}
    \centering
    \includegraphics[width=1\linewidth]{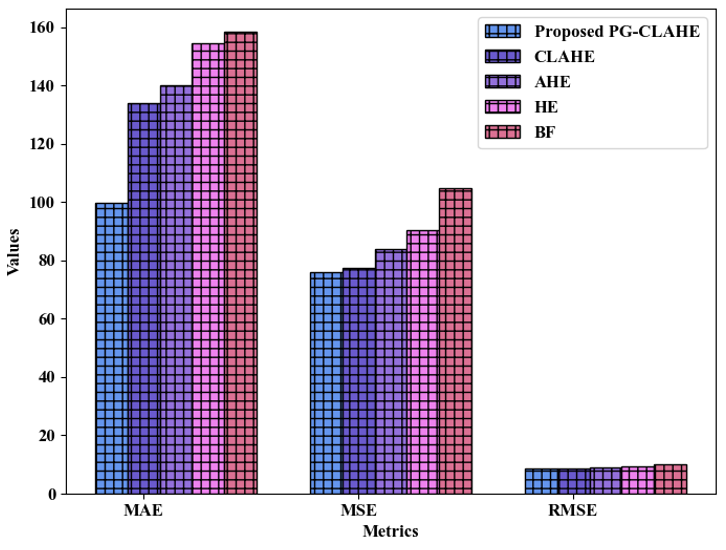}
    \caption{Analysis of MAE, MSE, and RMSE.}
    \label{fig:enter-label}
\end{figure}

Figure 3 illustrates the Mean Absolute Error (MAE), Mean Squared Error (MSE), and Root Mean Squared Error (RMSE) for both proposed and existing techniques. The proposed PG-CLAHE attained lower errors with an MAE of 99.87370556, MSE of 76.0186875, and RMSE of 8.718869623. In comparison, existing techniques, including CLAHE, AHE, HE, and BF, had higher errors and attained an average MAE of 146.6632203, MSE of 89.10923906, and RMSE of 9.424607049. The reduced errors in PG-CLAHE result from the Pareto Gini distribution technique for clipping limit adjustment. This enhances contrast more effectively than existing techniques.

\subsubsection{Performance Analysis of Object Segmentation}
The performance of the proposed E-YOLO and the existing techniques are analyzed and compared in terms of Mean Average Precision (MAP), Average Precision (AvP), and IOU.

\begin{table}[h]
    \centering
    \begin{tabular}{lllrrr}
        \toprule
        Techniques & MAP & AvP \\
        \midrule
        Proposed E-YOLO & 0.94563884 & 0.943820261 \\
        YOLO & 0.899554092 & 0.917271493 \\
        FRCNN & 0.899459486 & 0.916879817 \\
        SSD & 0.8855592782 & 0.886889476 \\
        VJ & 0.857678542 & 0.861389579 \\
        \bottomrule
    \end{tabular}
    \caption{MAP and AvP Analysis.}
\end{table}

Table 1 shows the MAP and AvP performance analysis of the proposed E-YOLO and the existing techniques like YOLO, Faster Region-based CNN (FRCNN), Single Shot MultiBox Detector (SSD), and Viola-Jones (VJ). The proposed E-YOLO incorporates the Easom function with overlapping elimination to identify small objects. As a result, it achieved a MAP of 0.94563884 and an AvP of 0.943820261. In contrast, the existing YOLO, FRCNN, SSD, and VJ achieved MAP between 0.857678542 and 0.899554092 and AvP between 0.861389579 and 0.917271493. The higher MAP and AvP of E-YOLO demonstrate its superior object segmentation performance.

\begin{figure}
    \centering
    \includegraphics[width=1\linewidth]{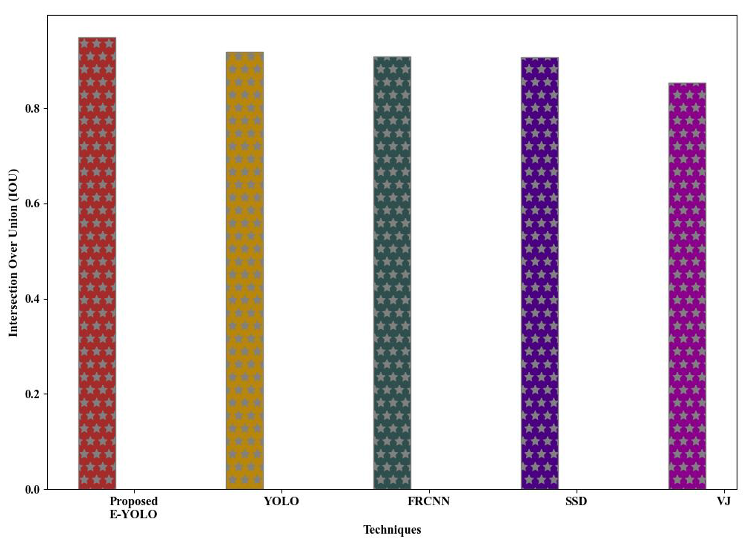}
    \caption{IOU Analysis}
    \label{fig:enter-label}
\end{figure}

Figure 4 illustrates the IOU performance of the proposed E-YOLO and existing techniques. The proposed E-YOLO achieved an IOU of 0.948156196, with a high value indicating its strong performance. In comparison, the existing models like YOLO, FRCNN, SSD, and VJ reached IOU values of 0.918258453, 0.90875172, 0.907456583, and 0.852800155, respectively, showing comparatively lower performance. The higher IOU of E-YOLO highlights its superior efficacy in object segmentation.

\subsubsection{Performance Analysis of Knowledge Graph Construction}
Here, the performance analysis of the proposed CRKG and the existing techniques based on Graph Generation Time (GGT) is shown in Figure 5.

\begin{figure}
    \centering
    \includegraphics[width=1\linewidth]{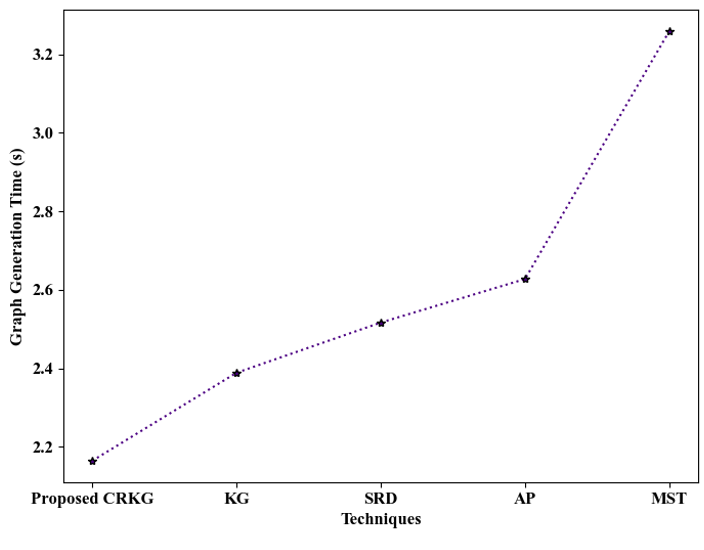}
    \caption{Analysis of Graph Generation Time}
    \label{fig:enter-label}
\end{figure}

The proposed CRKG utilized the Conditional Random technique to eliminate biases present in the image and achieved a GGT of 2.079750538s. In comparison, the existing KG, Spatial Relationship Detection (SRD), Affinity propagation (AP), and Minimum Spanning Tree (MST) took GGT of 2.203727722s, 2.406022787s, 2.506807089s, and 2.607764006s, correspondingly. This demonstrates that CRKG outperforms existing techniques in graph generation efficiency.

\subsubsection{Performance Analysis of Cross-Modal Query Understanding System}
The proposed CAZSSCL-MPGPT is compared and analyzed with the existing techniques, such as GPT, BERT, Bidirectional and Auto-Regressive Transformer (BART), and T5 using the COCO Dataset 2017 and vqav2-val Dataset. 

\textbf{COCO Dataset 2017}
In Figure 6, the performance analysis of the proposed CAZSSCL-MPGPT and existing techniques is validated on the COCO dataset 2017 in terms of F1-score and specificity.

\begin{figure}
    \centering
    \includegraphics[width=1\linewidth]{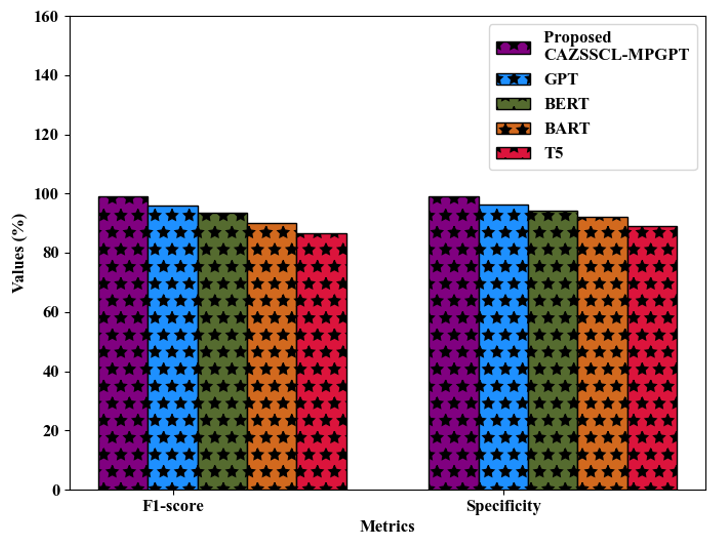}
    \caption{Performance Analysis of F1-Score and Specificity}
    \label{fig:enter-label}
\end{figure}

The proposed CAZSSCL-MPGPT achieved an F1-score of 99.14277995\% and a specificity of 99.15805022\%, outperforming the existing models, such as GPT, BERT, BART, and T5, which attained average F1-score and specificity values of 91.77962625\% and 91.75406204\%, respectively. This improvement is due to the use of mixup regularization that selects relevant features, and the Phish activation function boosts efficiency. Cross-attention and zero-shot learning align modalities and classify unseen data. Thus, the proposed CAZSSCL-MPGPT demonstrates superior performance in caption generation.

\begin{table}[h]
    \centering
    \begin{tabular}{lllrrr}
        \toprule
        Techniques &	Accuracy (\%) &	Precision (\%)	& BLEU & 	METEOR \\
Proposed CAZSSCL-MPGPT	& 99.14187362 &	99.15982902	& 0.991598 &	0.991428 \\
GPT &	96.05851979	& 96.09732121 &	0.960973 &	0.960619 \\
BERT &	93.62183428 &	93.5782967 &	0.935783 &	0.936334 \\
BART &	90.23604623 &	90.2381537 &	0.902382 &	0.902492 \\
T5	& 87.16006885 &	87.18679375 &	0.871868 &	0.871739 \\
        \bottomrule
    \end{tabular}
    \caption{Accuracy and Precision, BLEU, and METEOR Analysis.}
\end{table}

Table 2 shows the accuracy, precision, Bilingual Evaluation Understudy (BLEU), and Metric for Evaluation of Translation with Explicit Ordering (METEOR) analysis of the proposed CAZSSCL-MPGPT and existing techniques. The proposed CAZSSCL-MPGPT achieved an accuracy of 99.14187362\% and a precision of 99.15982902\%. In contrast, existing techniques, such as GPT, BERT, BART, and T5 demonstrated lower performance in both accuracy and precision. Likewise, the proposed CAZSSCL-MPGPT achieved BLEU and METEOR of 0.991598 and 0.991428, respectively. However, the existing GPT attained a BLEU of 0.960973 and T5 attained a METEOR of 0.871739. This demonstrates the superior performance of CAZSSCL-MPGPT in generating captions using the COCO dataset 2017.

\textbf{vqav2-val Dataset}
Figure 7 and Table 3 demonstrate the performance of the proposed CAZSSCL-MPGPT and the existing techniques in terms of False Positive Rate (FPR), False Negative Rate (FNR), accuracy, precision, BLEU, and METEOR for the vqav2-val Dataset.

\begin{figure}
    \centering
    \includegraphics[width=1\linewidth]{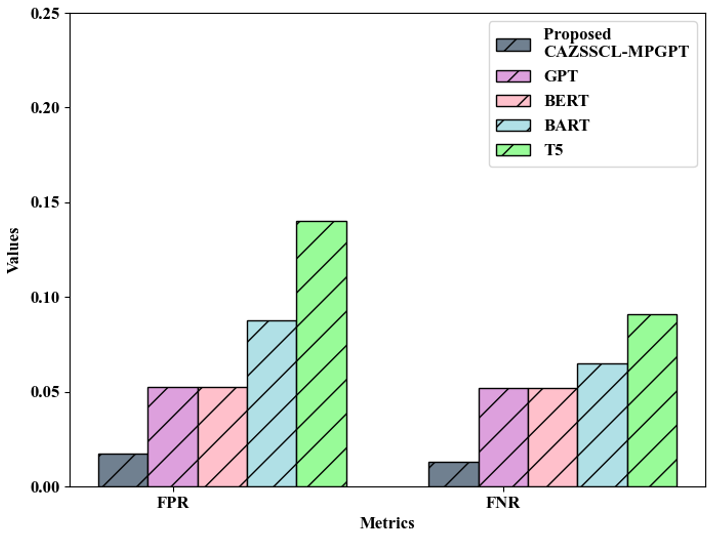}
    \caption{Analysis of FPR and FNR }
    \label{fig:enter-label}
\end{figure}

\begin{table}[h]
    \centering
    \begin{tabular}{lllrrr}
        \toprule
Techniques	& Accuracy (\%) &	Precision (\%)	& BLEU &	METEOR \\
Proposed CAZSSCL-MPGPT	& 98.43224393 &	98.39792963	& 0.983979296 &	0.984343216 \\
GPT	& 94.56531803 &	94.46905377 &	0.944690538 &	0.945766992 \\
BERT &	94.56531803	 & 94.46905377&	0.944690538&	0.945766992\\
BART&	0.919205012	&0.918353535&	0.918353535	&0.919371728\\
T5	&88.56587353&	88.44168613&	0.884416861&	0.885968422\\

        \bottomrule
    \end{tabular}
    \caption{Analysis of Accuracy, Precision, BLEU, and METEOR.}
\end{table}

The proposed CAZSSCL-MPGPT achieved a low FPR of 0.016063265 and a low FNR of 0.015292594, while existing techniques like GPT, BERT, BART, and T5 had higher average FPR and FNR values of 0.077319906 and 0.074597644, respectively. Moreover, the proposed CAZSSCL-MPGPT attained the accuracy, precision, BLEU, and METEOR of 98.43224393\%, 98.39792963\%, 0.983979296, and 0.984343216, respectively, outperforming the existing techniques. In contrast, the existing techniques like GPT, BERT, BART, and T5 achieved lower average accuracy, precision, BLEU, and METEOR of 69.65392865\%, 69.5745368\%, 0.923037868, and 0.924218533, respectively. This clearly demonstrates that the proposed CAZSSCL-MPGPT outperforms caption generation over the existing techniques.

\subsection{Comparative Analysis}
The efficacy of the proposed system is demonstrated by a comparison with related works.

\begin{table}[h]
    \centering
    \begin{tabular}{lllrrr}
        \toprule
        Author’s Name&	Technique/Method Used&	METEOR&	Drawbacks\\
Proposed Framework	&CAZSSCL-MPGPT&	0.991428&	The model didn’t concentrate on domain-specific contexts.\\
(Nursikuwagus et al., 2024)& SNN and LSTM&	0.670&	Ineffective for non-linearly separable data.\\
(Xiang et al., 2023)&	EVSD&	25.8&	Potential bias affected model generalizability.\\
(Im \& Chan, 2023)&	CNN-to-Bi-CARU	&31.23&	Lack of additional feature incorporation.\\
(Duhayyim et al., 2022)&	BiGRU&	30.00&	Bi-GRUs required significant computational resources.\\
(Iwamura et al., 2021)	&Motion-CNN	&26.7&	Object detection failures degraded performance.\\

        \bottomrule
    \end{tabular}
    \caption{Comparative Analysis with Related Works.}
\end{table}

In Table 4, the comparative analysis of the proposed CAZSSCL-MPGPT framework with existing techniques is evaluated on the COCO dataset. The existing techniques, such as the hybrid SNN with LSTM and EVSD have METEOR scores of 0.670 and 25.8, respectively. Also, they face challenges in providing accurate captions due to inefficiencies and biases. The CNN-to-Bi-CARU model and BiGRU-based systems, with METEOR scores of 31.23 and 30.00, respectively, struggle to fully capture contextual relationships between image content and textual descriptions. Models like Motion-CNN also face difficulty in object detection, resulting in a METEOR score of 26.7. In contrast, the proposed CAZSSCL-MPGPT framework outperforms these techniques with a METEOR score of 0.991428 on the COCO dataset. This higher score confirms its effectiveness, making it a strong choice for cross-modal query understanding.

\section{Conclusion}
An efficient LLM for the cross-modal query understanding system using DL-KeyBERT-based CAZSSCL-MPGPT is proposed in this framework. The experimental results show that the proposed CRKG achieves a GGT of 2.079750538s for knowledge graph construction. Similarly, the proposed PG-CLAHE achieved MAE and MSE values of 99.87370556 and 76.0186875, respectively. Additionally, the proposed E-YOLO segmented objects with an IOU of 0.948156196. The proposed CAZSSCL-MPGPT demonstrated outstanding performance in caption generation and VQA on both the COCO dataset 2017 and the vqav2-val dataset. Specifically, it achieved an accuracy of 99.14187362\%, BLEU of 0.983979296, and METEOR of 0.984343216 for the COCO 2017 dataset. Then, for the vqav2-val dataset, the model attained an accuracy of 98.43224393\% with BLEU scores of 0.983979296 and METEOR scores of 0.984343216. These results underscore the high performance and robustness of the proposed system in both caption generation and visual question-answering tasks.

\textbf{Future Scope}
Although the proposed cross-modal query understanding system generates captions efficiently, it does not focus on domain-specific contexts. In the future, the system can be enhanced by concentrating on domain-based datasets, such as healthcare, education, travel, and hospitality, along with integrating cultural events and polarity considerations.

\end{document}